# Transformer-Lite: High-efficiency Deployment of Large Language Models on Mobile Phone GPUs

Luchang Li*   Sheng Qian   Jie Lu   Lunxi Yuan   Rui Wang   Qin Xie

OPPO AI Center

## Abstract

The Large Language Model (LLM) is widely employed for tasks such as intelligent assistants, text summarization, translation, and multi-modality on mobile phones. However, the current methods for on-device LLM deployment maintain slow inference speed, which causes poor user experience. To facilitate high-efficiency LLM deployment on device GPUs, we propose four optimization techniques: (a) a symbolic expression-based approach to support dynamic shape model inference; (b) operator optimizations and execution priority setting to enhance inference speed and reduce phone lagging; (c) an FP4 quantization method termed E0M4 to reduce dequantization overhead; (d) a sub-tensor-based technique to eliminate the need for copying KV cache after LLM inference. Furthermore, we implement these methods in our mobile inference engine, Transformer-Lite, which is compatible with both Qualcomm and MTK processors. We evaluated Transformer-Lite's performance using LLMs with varied architectures and parameters ranging from 2B to 14B. Specifically, we achieved prefill and decoding speeds of 121 token/s and 14 token/s for ChatGLM2 6B, and 330 token/s and 30 token/s for smaller Gemma 2B, respectively. Compared with CPU-based FastLLM and GPU-based MLC-LLM, our engine attains over 10x speedup for the prefill speed and 2~3x speedup for the decoding speed.

## 1 Introduction

The LLM has demonstrated superior performance in a range of tasks[1, 20, 32, 39], proving versatile in applications such as intelligent assistants, text summarization, translation, and multi-modality tasks on mobile devices. Although the performance is excellent, the deployment of LLMs requires substantial computing power and memory bandwidth. Hence, current applications predominantly adopt the cloud-based deployment. Given the substantial cost involved in the cloud-based deployment of LLMs and the continuous advancements in the performance of mobile devices, native deployment of LLMs on mobile phones not only curtails the high costs associated with cloud deployment but also broadens the prospective applications of LLMs on mobile phones. Therefore, deploying LLMs on devices is becoming a research hotspot.

Compared with cloud servers, deploying LLMs on mobile phones is constrained by limited hardware performance, memory bandwidth, and storage capacity. As a result, only smaller parameter models can be employed, and the parameter range of current mainstream LLMs is approximately from 1B to 13B [24, 44–47, 47]. The speed of LLM inference is critical for user experience. Although models with reduced parameters operate at a faster inference speed, their accuracy is inevitably compromised. To enhance the user experience, high-efficiency LLM deployment that secures both high accuracy and efficient inference is essential.

*Corresponding author: liluchang@oppo.com

To deploy LLM on device, currently, two types of methods can be used: generic mobile inference engines such as TFLite[48], MNN[18], NCNN[49], etc., or LLM-specific engines including Llama.cpp[50], MLC-LLM[51], FastLLM[52], etc. Conventionally, computer vision (CV) models were widely deployed on devices using generic mobile inference engines[2, 8, 33], such as ResNet[15], MobileNet[17], etc. The advantage of these engines lies in their ability to directly use serialized models exported from training frameworks, such as ONNX[53], thereby eliminating the need for users to re-describe the model structure and simplifying support for various model architectures. However, these engines are primarily optimized for static shape CV models. LLMs substantially differ from CV models in terms of dynamic input shapes, model structures, operator types, tensor dimensions, etc. Consequently, optimizations for CV models are typically not directly applicable to LLMs. As a result, high-efficient LLM deployment based on device GPUs is usually not supported.

In contrast, LLM-specific engines are specifically designed for transformer-based LLM deployment and demonstrate the feasibility of deploying 1B-7B models on devices using CPUs or GPUs. However, these engines have not yet fully utilized the hardware performance and memory bandwidth, resulting in slow prefill and decoding speeds, which in turn degrade the user experience. Furthermore, these engines necessitate the re-descriptions of model structures via C++ or TVM script, complicating support for new model architectures. Notably, these LLM-specific engines do not accommodate other types of models such as CV and vision transformer (ViT).

To support LLM deployment on device GPUs both conveniently and efficiently, we advocate for combining the strengths of generic mobile inference engines and LLM-specific engines. To address this issue, first, to enable high-efficiency LLM deployment on device GPUs, we propose four optimization techniques:

- A symbolic expression-based approach to support dynamic shape model inference, including dynamic shape derivation, memory reuse, execution scheduling, etc.
- Operator optimizations and execution priority setting to enhance both performance and reduce phone lagging.
- An FP4 quantization method termed E0M4 minimizes performance overhead in dequantization, thereby enabling more efficient matrix multiplication.
- A sub-tensor-based approach to circumvent KV cache copying from model outputs to model inputs after each LLM inference iteration.

Additionally, we developed a novel mobile inference engine called Transformer-Lite and integrated these optimizations. Our engine deploys LLM using ONNX models exported by training frameworks such as PyTorch, thereby making LLM deployment convenient and enabling easy support for new model types.

The rest of the paper is organized as follows. Section 2 elaborates on the details of the aforementioned key optimizations. Section 3 includes an evaluation of various LLM models using both Qualcomm and MediaTek processors. Section 4 revisits related work, and Section 5 concludes the paper.

# 2 Method

## 2.1 Symbolic expression-based dynamic shape inference

In this section, we first describe the basics of dynamic shape inference and propose a symbolic expression-based method to express and infer the dynamic shape of tensors. Next, we describe how we perform memory reuse between dynamic shape tensors. Finally, we describe the methods for reducing the time consumption of shape updating during LLM inference.

**Basics of dynamic shape inference**

In the past, CV models were widely deployed on devices, such as ResNet, MobileNet, etc. These models typically feature fixed input shapes, enabling the unique determination of all tensor shapes within the models. We refer to this scenario as static shape inference. However, in LLM, the newly generated token from each iteration is added to the input token sequence as the input for the next iteration. Therefore, LLM is a dynamic shape input scenario, where the input shape for each iteration changes. This leads to changes in the shape of some activation tensors in the model, which pose significant challenges in memory reuse and operators' performance optimizations.

**Represent unknown dimensions by symbolic expression**

Certain model serialization methods and inference engines currently utilize specialized techniques to signify unascertained dimensions of shapes. For instance, the ONNX typically uses -1 or string representations to embody unknown dimensions of dynamic shapes, exemplified by [-1, 128] or ["batch", 128]. TVM[4] also endorses the use of symbols for annotating the relationship of shapes between dynamic tensors. With inspiration drawn from ONNX and TVM, the hypothesis is posited that symbolic expressions are essential for delineating the ambiguous dimensions of dynamic shapes to precisely determine the relation among tensors. These symbols are determined as valid strings adhering to the naming convention of C language variables, where the symbolic expression is a mathematical computation involving symbols and integers, such as "batch" * 16. Such symbolic expressions enable us to accurately resolve the shape relationship amid tensors, proving critical for memory reuse and performance optimization.

**Dynamic shape derivation and execution**

Although ONNX permits the use of strings to represent unknown tensor dimensions, it does not provide comprehensive dynamic shape derivation. For example, take a Reshape operator with an input shape of [1, "N", 4096] and a target shape of [1, -1, 32, 128]. The resultant output shape could be [1, "unk_1", 32, 128], which obscures the size relationship between the input and output tensors. To rectify this, we introduce comprehensive shape derivation based on symbolic expressions. For instance, concatenating shape ["sumN-N",1,2,128] with ["N",1,2,128] at axis 0 yields an output shape of ["sumN",1,2,128]. To this end, we utilize a method akin to Nimble[31] and DISC[43] to classify operators in deep learning models into two categories: tensor computing operators and shape computing operators. The latter category encapsulates the Shape operators and those whose inputs depend on the Shape operators' results. Shape computing operators are executed on the CPU and calculate shape information. Conversely, tensor computing operators, responsible for computing activation tensors and weights, are executed on the GPU. Figure 1 provides an example of operator classification for a demonstration model.

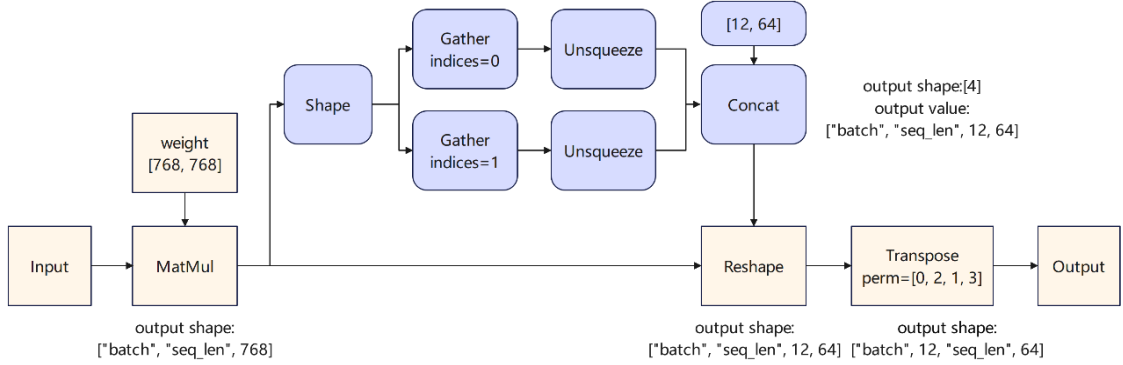

Figure 1. Example of shape computing operator and tensor computing operator classification. The blue boxes represent the shape computing operators and the yellow boxes represent the tensor computing operators.

The classification yields advantages in both the shape derivation and model inference stages. First, during the shape derivation stage, when deducing output shapes based on input shapes for each operator, executing the inferred operator is necessary if it is a shape computing operator. Python's SymPy[54] tool is employed to facilitate symbolic expression computation throughout this process. Once shape derivation is completed, the symbolic expressions of each tensor are stored. Before the model inference stage, the actual shapes of dynamic shape tensors are computed, according to the inferred symbolic expressions and the user-defined integer value for each unknown symbol. This method eliminates the need for time-consuming shape derivation in the model inference stage.

Second, in the model inference stage, all shape computing operators are executed at once using the CPU, followed by the execution of all tensor computing operators on the GPU. This approach minimizes frequent CPU-GPU synchronization. Performance would suffer significantly if operators were executed solely based on topological order, as this would necessitate numerous synchronizations due to the boundary between GPU and CPU. Moreover, during inference, the shape computing operators between the Shape and Reshape operators need not be executed again, as the symbolic expression of the reshape output shape has already been obtained. Consequently, the number of shape computing operators requiring execution typically reduces to zero during the inference.

**Memory reuse**

During the inference stage, each activation tensor maintains a finite lifecycle. However, the continuous allocation and release of memory in each tensor can lead to considerable performance overhead. Memory reuse of base tensors by other tensors with identical or smaller memory requirements can avoid this performance overhead[18]. The prerequisite to implementing memory reuse is acquiring the memory size relationship between tensors, which is straightforward in static shape inference. Symbolic expressions facilitate the easy determination of this relationship. First, the memory size of each tensor is calculated as the number of elements times the number of bytes of the data type, creating a symbolic expression for each tensor's memory size. Subtraction and division are then combined to discern the memory size relationship between tensors. For instance, dividing "N" * 4096 by "N" * 32 * 128 results in an integer of 1, indicating equal size and the possibility of memory reuse. On the other hand, when "N" * 4096 is compared to "sumN" * 2 * 128, both subtraction and division yield another symbolic expression instead of an established integer. Therefore, their size relationship is undefined unless all potential input shapes are explored, which means memory reuse cannot be performed.

To facilitate memory reuse, the OpenCL buffer memory type is employed to allocate all activation tensors. As for weights, they are allocated using either the OpenCL buffer or image memory type based on the prerequisites of the consuming operators. Operators such as matrix multiplication may necessitate using an OpenCL image as activation input for optimum performance. Therefore, we utilize the image from buffer extension[55] to generate an image reference from the buffer, bypassing the necessity for converting data from diverse memory types or allocating new memory.

**Reduce shape updating time consumption**

Typically, each inference iteration of LLM necessitates updating the input shapes. During the shape updating process, it is essential to recompute the dynamic shape tensor's shape and update the parameters of operators with dynamic shape input or output tensors. Consequently, frequent input shape updates often result in a non-negligible performance overhead. To circumvent this issue, we employ the attention mask mechanism to pad the model input sequence length to multiples of 64 or 128 during the decoding stage. This approach allows us to update the shape only when the actual token count surpasses the padding length, rendering the shape updating time overhead for each inference iteration negligible. Furthermore, based on the symbolic expression of the activation tensors' shape, we precisely allocate the memory required for maximum sequence length inference at the onset of the inference, thus eliminating the necessity for memory reallocation during shape updating.

## 2.2 Operator and lagging optimizations

**Operator optimizations.** The LLMs typically harness 4-bit quantization to diminish the model size and memory bandwidth demands[12, 22]. Despite this, to maintain accuracy, activation computations are generally carried out using half-precision. Conventional BLAS libraries do not accommodate matrix multiplication that incorporates both half-precision activation and 4-bit quantized weight. Therefore, to maximize the memory bandwidth advantages offered by 4-bit quantized weights, we implement matrix multiplications that directly manage the half-precision activation and 4-bit quantized weight. Furthermore, the prefill and decoding stages exhibit different shape scenarios for matrix multiplications. In the prefill stage, matrix multiplications conduct matrix-matrix computations, while in the decoding stage, these operators are reduced to vector-matrix computations. As a result, we devise two distinct matrix multiplication implementations tailored specifically for prefill and decoding stages.

Although matrix multiplications represent the most time-consuming operators in LLM inference, other operators still contribute to a considerable overhead. Consequently, we conduct intensive operator fusions to mitigate the time consumption of these operators. For instance, we fuse the smaller operators that compose layer-norm and rms-norm. Additionally, we automatically fuse the elementary-wise operator with other operators.

**lagging optimizations.** In addition to facilitating deep learning inference, GPUs also perform graphics rendering on mobile devices. Executing inference on large models with substantial computational demands and extended inference durations may result in considerable lagging. To address this issue, we leverage the OpenCL extensions provided by Qualcomm and ARM[56, 57], setting the execution priority of operators of deep learning models to the lowest level. This approach effectively alleviates phone lagging.

## 2.3 E0M4 FP4 quantization

**Background.** Quantization methods such as GPTQ[12] and AWQ[22] quantize LLM weights from floating point to 4-bit integers through group-wise manners. However, to maintain high accuracy, activations typically employ higher-precision computations, such as half precision. Consequently, during model inference, the weights need to be dequantized to floating point to perform matrix multiplication. However, this dequantization process necessitates the explicit conversion of the weights from integer to floating point. According to the IEEE-754[58] standard, the binary of floating-point data is comprised of three parts: sign, exponent, and fraction. An illustration of the binary storage of FP16 is provided in Figure 2[59]. During the conversion of an integer to a floating-point number, intricate algorithms or instructions are utilized to determine the value of these three parts. Hence, the data type conversion in dequantization triggers a non-negligible performance overhead.

**Quantization.** To address this issue, we propose an FP4 quantization method termed E0M4, which means zero bits for exponent, and four bits for fraction. In the dequantization stage, the 4-bit quantized data can be converted to floating point by executing only two bitwise operations. We adopt group-wise quantization and convert the tensor of each subgroup (represented as w0) into a fresh floating-point tensor (denoted as w1) via scaling and bias coefficients. We ensure that all w1 elements fall within the range $[2^n, 2^{n+1} - eps]$. Here, 'n' is an integer, usually selected as 1 or 2, and 'eps' is a minor floating-point coefficient. The binary representation of all floating-point elements in w1 shares the same sign bit and exponent bits but varies in the fraction bits. Therefore, we isolate the highest four bits of the fraction bits (represented as y1, as depicted in Figure 2) to serve as the quantization result. To augment the accuracy of quantization, we isolate an additional one bit (represented as y2, as illustrated in Figure 2) for rounding: y1 = min (y1+y2, 15). Here, 15 is the maximum value expressible by a 4-bit unsigned integer. It is important to zero the non-intercepted fraction bits of the bias coefficient to ensure the accurate quantization of the zero elements.

**Dequantization.** We execute a left shift on the 4-bit quantization result y1, and subsequently carry out a bitwise OR operation with the constant sign and exponent bits to derive the binary representation of the floating-point number:

w1_half = as_half (CONST_EXP_BIN_PART | (w1_fp4 << CONST_BIT_SHIFT_NUM));

The as_half function is an OpenCL method, which reinterprets binary without necessitating additional computations. Theoretically, our E0M4 quantization can diminish the performance overhead in the dequantization process and seamlessly integrate with quantization methods such as GPTQ and AWQ. Additionally, our approach can be employed for other bit-width quantization.

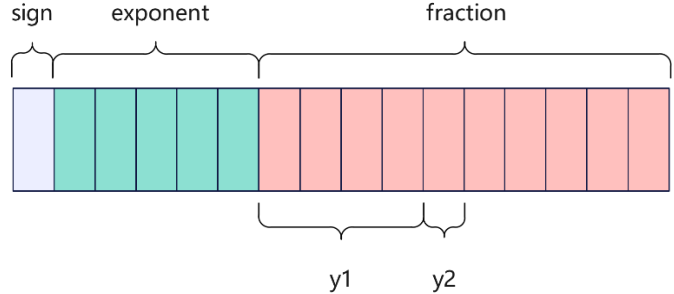

Figure 2. Demonstration of FP16 binary representation and E0M4 FP4 quantization. The y1 and y2 are isolated bits for quantization results and rounding respectively.

## 2.3 KV cache copy optimization

**Background**. KV cache, a critical component in LLM inference acceleration, effectively avoids unnecessary recalculation of earlier tokens. During the LLM inference, the KV caches, created by the preceding token, operate as model inputs. These are then concatenated with the fresh caches generated by the new input tokens, and the resultant concatenated KV caches serve as model outputs. This procedure encounters two principal complications. First, the model inputs and outputs store the KV cache twice, consuming a significant memory volume, despite their content being primarily identical. Second, in the subsequent iteration, the output KV cache must be copied to the model inputs, thereby causing a decrease in inference speed.

**Optimization of KV cache copying**. We improve the KV cache storage, ensuring that only a single instance of the KV cache tensor requires storage, and employed sub-tensor technology to circumvent supplementary copying from output caches to input caches. Initially, we alter the ONNX model to exclusively output the recently generated cache, instead of concatenated caches. Subsequently, as depicted in Figure 3, we create a suitably large tensor for each input KV cache, contingent on the maximum sequence length needed for the inference task. We then generate sub-tensors based on varying address offsets and utilize these sub-tensors as model input and output KV caches. This enables the new KV caches to be directly written to the appropriate positions during model inference, consequently preventing additional copying processes.

To eliminate the time overhead involved in creating sub-tensors, a viable solution involves substituting the concatenation operator with a custom in-place operator, explicitly designated for KV cache concatenation. This operator employs the identical KV cache tensor for both input and output and incorporates position_ids as supplementary input to ascertain the sub-tensor address.

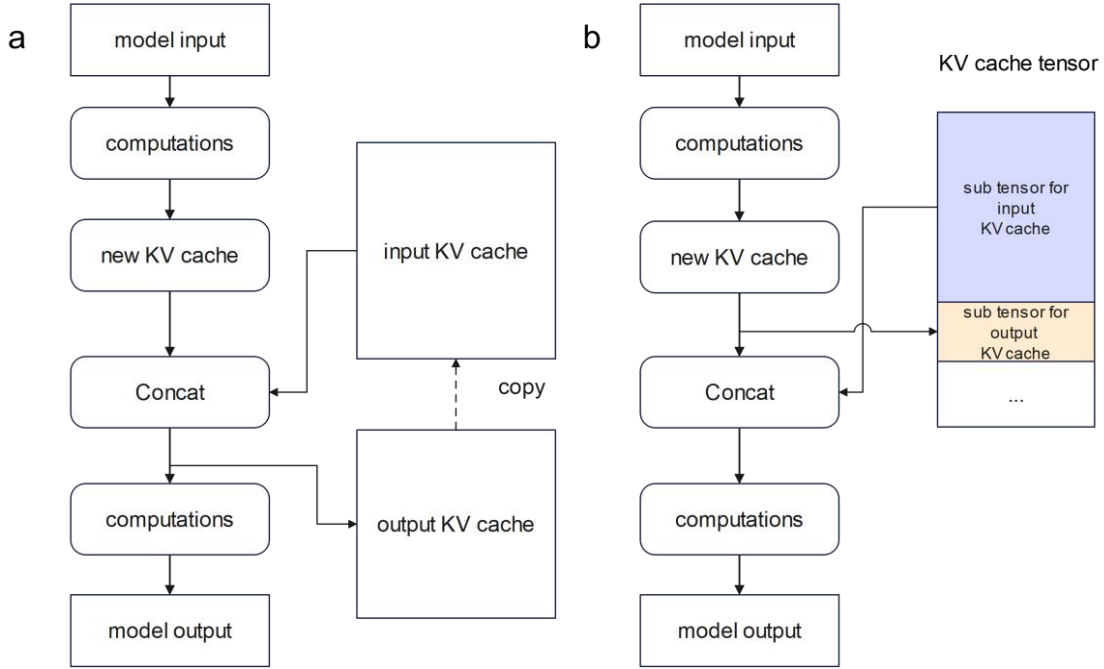

Figure 3. Diagram of KV cache copying optimization. (a) The original LLM KV cache management with duplicative cache storage and cache replication. (b) The optimized KV cache management with singular storage and no KV cache copying.

**Modifications to KV cache format**. Our approach calls for a special data format for the KV cache. Specifically, the variable sequence length dimension of the KV cache must occupy the first dimension that is not 1. For instance, when the batch is 1, the KV cache format for ChatGLM2 6B is ["seq_len", 1, 2, 128], a format that satisfies our requirement. Conversely, for Llama2 7B, the KV cache format is [1, 32, "seq_len", 128], a format that does not fulfill our requirements. Nevertheless, we can incorporate transpose operators to modify Llama's KV cache into [1, "seq_len", 32, 128] to make it align with our requirements. The insertion of excessive transpose operators might lead to performance overhead; hence, referencing ChatGLM2 to refine the KV cache using more efficient techniques can potentially enhance performance.

# 3 Experiments

## 3.1 Settings

**Devices.** We selected two mobile phones for testing: the OPPO Find X7 24GB memory version, equipped with an MTK Dimensity 9300 processor featuring an ARM Mali-G720 Immortalis MC12 GPU, and the OPPO Find X7 Ultra 12GB memory version, equipped with a Qualcomm Snapdragon 8 Gen 3 processor featuring an Adreno 750 GPU.

**Models.** We chose five LLM models with varying structures and parameter sizes, namely Gemma 2B, Qwen1.5 4B, ChatGLM2 6B, Llama2 7B, and Qwen1.5 14B. We tested these models under three prompt lengths (128, 1024, and 2048) to evaluate their performance in different application scenarios, such as dialogue and summarization.

**Evaluations.** For our Transformer-Lite engine, we exported an ONNX model from each LLM for deployment[60]. To make the LLMs more efficient and avoid excessive shape computing operators, we made minor modifications to the PyTorch model before exporting LLM to ONNX. For instance, we directly used the expanded attention mask as model input instead of expanding the 2D attention mask during inference. In all evaluations, we did not switch the phone to high-performance mode nor did we use acceleration techniques such as sparsity[11, 25, 28, 35] or speculative decoding[3, 21, 27, 36, 40]. By default, all LLM models or matrix multiplications in Transformer-Lite evaluations are quantized by group-size quantization with a group size of 128.

**Baselines.** We compared the performance of GPU inference with MLC-LLM and CPU inference with FastLLM. We used the officially converted model weights provided by the two engines. A full comparison of these two engines with all models was not performed since only specific models were officially provided. For MLC-LLM, we tested it using its Android app and employed the prefill and decoding speed displayed by the app as performance results. For FastLLM, we compiled it from the source code and added time measurement to the source code to determine its inference speed.

### 3.2 Performance of Transformer-Lite

The evaluation of Transformer-Lite engine for different models under different devices and prompt lengths is shown in Figure 4.

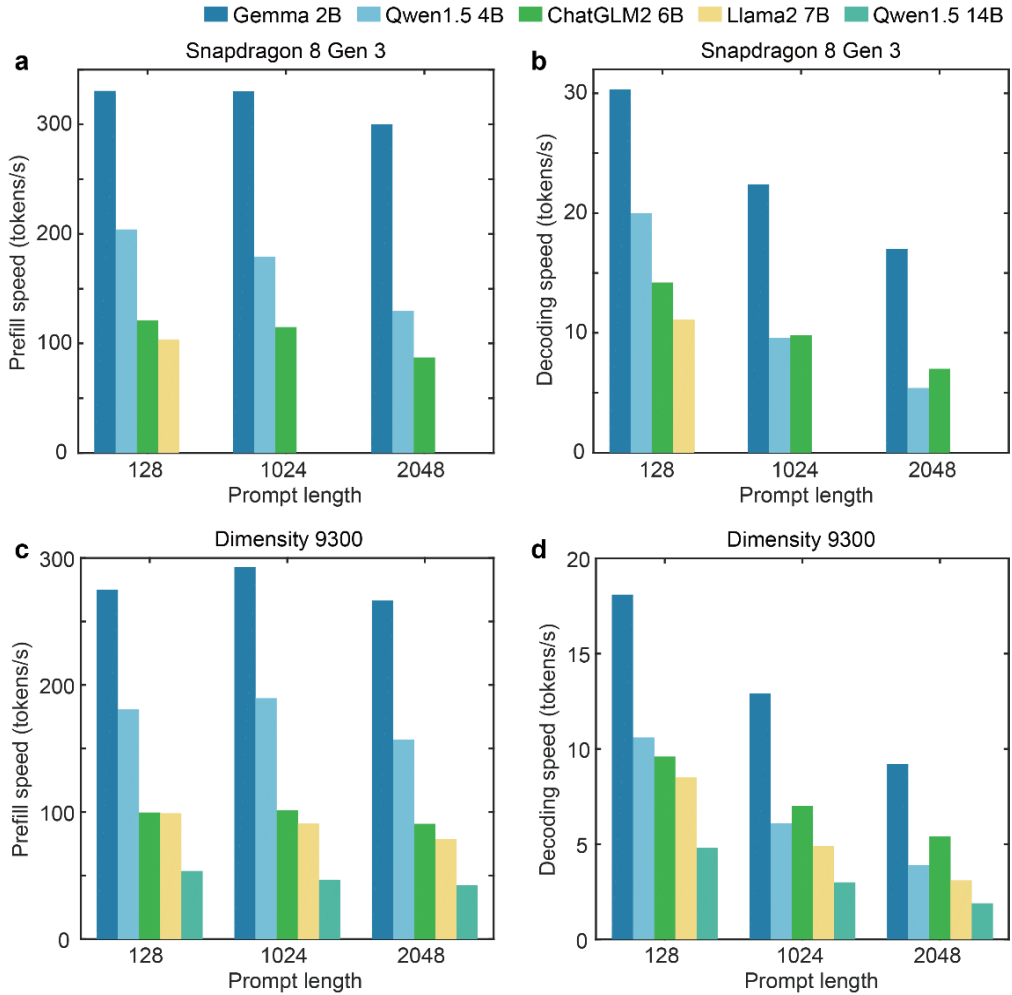

Figure 4. Prefill and decoding speed evaluation of Transformer-Lite. (a-b) Prefill and decoding speeds of Transformer-Lite based on Qualcomm Snapdragon 8 gen 3 processor respectively. (c-d) Prefill and decoding speeds of Transformer-Lite based on MTK Dimensity 9300 processor respectively.

**Prefill speed.** The prefill speed exhibits minimal sensitivity to prompt lengths, typically achieving optimum performance at a length of 1024, as shown in Figure 4 (a, c). This occurrence is due to larger shapes' potential to fully utilize the hardware performance. Nonetheless, as the prompt length expands, the time consumption of attention computation rises significantly, leading to a decline in prefill speed at longer prompt lengths. Furthermore, as anticipated, both the prefill and decoding speeds are nearly inversely proportional to the model parameter size. As a result, smaller models contribute to higher speeds, and the performance of models with different parameter sizes can be roughly estimated.

**Decoding speed.** A consistent decrease in decoding speed is observed as prompt length increases, as shown in Figure 4 (b, d). This phenomenon is attributed to the increased time consumption of the attention computation, whereas the time consumption of matrix multiplications is constant. Additionally, the decoding speed decreases more rapidly for Llama2 7B and Qwen1.5 4B compared with ChatGLM2 6B and Gemma 2B, because transpose operators were incorporated to modify the KV cache format, subsequently leading to additional time consumption. In contrast, ChatGLM2 6B and Gemma 2B do not necessitate any alterations to the KV cache format. It should be feasible to identify improved methods of modifying the KV cache format that reduce the performance overhead. The implementation of flash attention[5, 6] or flash decoding[7] significantly mitigates the decrease in decoding speed for server inference, proposing that the application of flash attention in mobile devices is an area of future work. Furthermore, the utilization of new LLM model architectures like RWKV[29] and Mamba[14] can also prevent a decline in decoding performance due to increased prompt length.

**Comparison of performance between two processor types.** Usually, we observe higher performance on the Adreno GPU in the Qualcomm processor, whereas the performance on the ARM GPU in the MTK processor tends to be slightly slower. These distinctions arise due to the different architectures of these two GPU types, necessitating the use of differing matrix multiplication implementations. Presently, as we employ identical matrix multiplication implementations, the adoption of optimizations tailored to individual GPU architectures to achieve peak performance is an area for future work.

**Maximum deployable model size.** Based on a phone with 12GB memory, as shown in Figure 4 (a, b), we facilitate 7B model inference with a maximum of 900 tokens for inference length, and increased token length inference would necessitate more memory. This is a consequence of the substantial increase in KV cache memory usage with the token length's augmentation. At present, we have not employed KV cache quantization[10, 13, 16, 37], which could considerably cut down the KV cache memory usage and thus support a longer token inference. On a phone with 24GB memory, we demonstrate the successful deployment of the Qwen1.5 14B model with a prompt size of 2048 (Figure 4 (c, d)).

**Performance gap relative to theoretical limits.** We examine whether the maximum speed has been realized for both the prefill and decoding stages. The prefill stage is constrained by the hardware performance characterized by the FLOPS[42]. At present, we have achieved 1.6 TFLOPS for half-precision matrix multiplication for the Snapdragon 8 gen3 processor. However, based on our evaluation with the ArchProbe[61] tool, this processor possesses a computation performance of 5.6 TFLOPS in half precision. Therefore, theoretically, there's still considerable room for enhancing prefill speed. Indeed, the freshest results from TFLite have showcased a prefill speed that's twice as fast for Gemma 2B[45]. For the decoding stage, the speed is restricted by memory bandwidth.[38] At shorter prompt lengths, we have attained speeds approaching the memory bandwidth limit for the Qualcomm processor; hence, significant improvements in decoding speed cannot be realized without the incorporation of other acceleration techniques such as sparsity[11, 25, 28] and speculative decoding[3, 21, 26, 27, 36, 40]. At longer prompt lengths, techniques like flash attention or refined model structures can mitigate the decreasing performance.

## 3.3 Comparison with MLC-LLM and FastLLM

We compare our Transformer-Lite engine with MLC-LLM using the Gemma 2B model based on GPU inference, as illustrated in Figure 5, and with FastLLM using the ChatGLM2 6B model for CPU inference, as demonstrated in Figure 6. Transformer-Lite typically achieves over 10x speedup for prefill speeds, and 2-3x speedup for decoding speed when compared with both MLC-LLM and FastLLM. Specifically, based on Snapdragon 8 gen 3 processer and a prompt length of 128, Transformer-Lite achieves prefill and decoding speeds of 330 token/s and 30 token/s for Gemma 2B, 121 token/s and 14 token/s for ChatGLM2 6B, respectively; In contrast, MLC-LLM achieves 25 token/s and 11 token/s for Gemma 2B, and FastLLM achieves 7 token/s and 1.2 token/s for ChatGLM2 6B respectively.

Moreover, we obtain similar prefill performance for Qualcomm Snapdragon 8 gen 3 and MTK Dimensity 9300 processors, whereas MLC-LLM exhibits a much slower prefill speed on the MTK processor. For example, based on Gemma 2B and a prompt length of 128, Transformer-Lite achieves 330 token/s and 275 token/s prefill speed for the two processors respectively. Whereas MLC-LLM achieves 25 token/s for the Qualcomm processor, which is five times the speed obtained for the MTK processor.

The enhanced performance is attributed to two reasons. First, the performance of the GPU is considerably faster than the CPU, making the GPU more appropriate for LLM inference. Second, we implemented superior operator optimizations to better utilize GPU performance and memory bandwidth in the prefill and decoding stages, respectively.

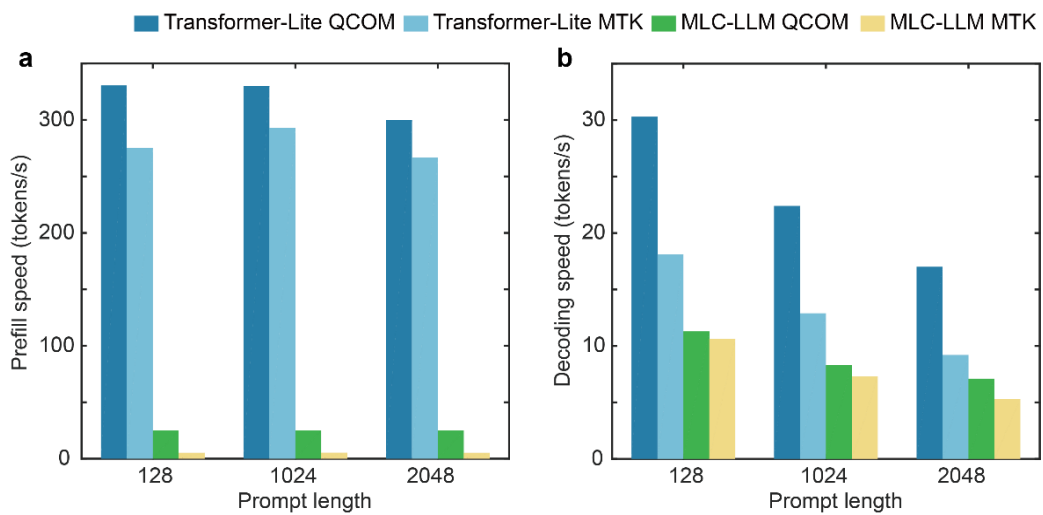

Figure 5. Comparison between Transformer-Lite and GPU-based MLC-LLM based on Gemma 2B model.

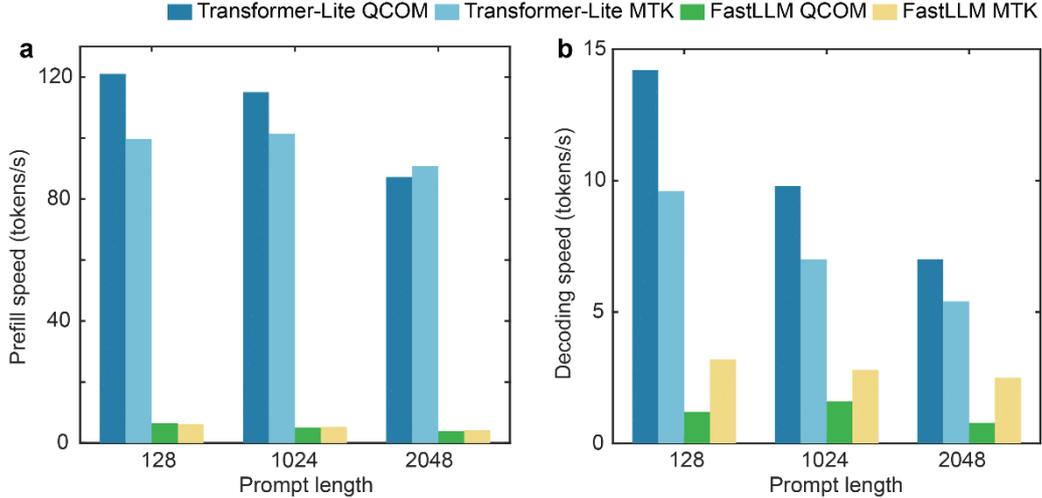

Figure 6. Comparison between Transformer-Lite and CPU-based FastLLM based on ChatGLM2 6B.

## 3.4 Evaluation of FP4 quantization

**Performance evaluation.** We examine the performance enhancement in matrix multiplication, achieved through the E0M4 FP4 dequantization in comparison to INT4 dequantization. We evaluate three matrix multiplication weight shapes and acquired the matrix multiplication latency by utilizing the OpenCL profiling interface. The results are presented in Table 1. For the Snapdragon 8 gen3 processor, the FP4 quantization method neither improves nor degrades the performance. However, for the Dimensity 9300 processor, the performance is improved by over 30%. It is important to note that the profiling time of the ARM GPU is typically longer than the actual latency. Therefore, the profiling time of the Adreno GPU in the Snapdragon 8 gen 3 processor and that of the ARM GPU in the MTK Dimensity 9300 should not be directly compared.

| Processor | K | N | Latency of FP4 (ms) | Latency of INT4 (ms) | Acceleration |
|---|---|---|---|---|---|
| Snapdragon 8 gen 3 | 4096 | 4096 | 0.16 | 0.16 | 1.00 |
| Snapdragon 8 gen 3 | 4096 | 11008 | 0.39 | 0.39 | 1.00 |
| Snapdragon 8 gen 3 | 11008 | 4096 | 0.42 | 0.42 | 1.00 |
| MTK Dimensity 9300 | 4096 | 4096 | 3.2 | 5.0 | 1.56 |
| MTK Dimensity 9300 | 4096 | 11008 | 6.0 | 8.3 | 1.38 |
| MTK Dimensity 9300 | 11008 | 4096 | 6.1 | 8.1 | 1.33 |

Table 1. Latency of matrix multiplication under different weight shapes and devices. The shape of weight is [K, N], while the activation tensor shape in all evaluations is [1, k]. Note that the latencies are acquired by the OpenCL profiling interface, and the latency of the two

types of processors should be directly compared.

**Accuracy evaluation.** We compare the accuracy of the FP4 quantization method with INT4 quantization by analyzing the mean absolute error (MAE) between the original and dequantized matrix multiplication weights. We examine all matrix multiplications within the first layer of Qwen1.5-4B-Chat, as detailed in Table 2. The results indicate that for all matrix multiplications, our FP4 quantization achieves approximately 4.5% smaller MAE compared with INT4 quantization. The enhanced accuracy can be attributed to our quantization's utilization of constant floating-point exponent information, which is not necessarily preserved in the quantized data.

| Matrix multiplication | MAE of FP4 | MAE of INT4 | Ratio |
|---|---|---|---|
| **q_proj** | 0.002815 | 0.002943 | 0.957 |
| **k_proj** | 0.002577 | 0.002699 | 0.955 |
| **v_proj** | 0.0009236 | 0.0009675 | 0.955 |
| **o_proj** | 0.0007606 | 0.0007963 | 0.955 |
| **up_proj** | 0.001328 | 0.001391 | 0.955 |
| **gate_proj** | 0.001307 | 0.001369 | 0.955 |
| **down_proj** | 0.001323 | 0.001386 | 0.955 |

Table 2. The MAE between the original and dequantized matrix multiplication weights in the first layers of Qwen1.5-4B-Chat.

## 3.5 Demonstration of ViT model deployment

Employing deployment based on the ONNX model format allows our Transformer-Lite engine to be model structure agnostic. This not only circumvents the need for model structure re-description within our engine but also facilitates the deployment of other types of deep learning models, including CV and ViT. We demonstrate this versatility with the OpenAI CLIP model[30, 62], with performance evaluation presented in Table 3. The results indicate that our engine can efficiently encode both images and text, potentially enabling high-performance image retrieval applications to be deployed on devices.

| Model | Processor | Latency (ms) |
|---|---|---|
| **image encoder** | Snapdragon 8 gen 3 | 19.2 |
| **image encoder** | MTK Dimensity 9300 | 29.5 |
| **text encoder** | Snapdragon 8 gen 3 | 13.9 |
| **text encoder** | MTK Dimensity 9300 | 27.2 |

Table 3. Performance of OpenAI CLIP clip-vit-base-patch32 image and text encoder based on two types of processors. The image encoder input shape is [1,3,224,224], and the text encoder input shape is [1,77], all models are executed by half precision.

# 4 Related Work

## 4.1 Deployment of LLM on device

Conventionally, mobile inference engines such as TFLite[48], MNN[18], and NCNN[49] have been employed for the deployment of deep learning models. Nevertheless, most of these engines exclusively grant support to static shape models on GPUs, with substantial optimization for CV models but insufficient optimization for transformer-based models. Attempts have been made by these engines to aid in deploying LLM on devices. For instance, MNN presently backs the deployment of LLM on CPUs, while TFLite has published performances for several 1B to 3B models deployed on both CPUs and GPUs[45]. Aside from these, there are LLM-specific engines including llama.cpp, FastLLM, and MLC-LLM[50–52], which predominantly support LLM inference on CPUs, though MLC-LLM also supports GPUs. Often, these LLM-specific engines necessitate the re-description of the LLM model structure in C++ languages or TVM script, rendering them inconvenient for new LLM structures and not supporting other model types such as CV and ViT models. Consequently, we merge the merits of traditional mobile inference engines and LLM-specific engines, deploying LLM via ONNX and employing exhaustive optimizations for the transformer-based models.

Apart from CPU and GPU, LLM can also be deployed on Neural Processing Unit (NPU)[47]. Compared to a GPU, the NPUs' performance is vastly superior, enabling much higher prefill speed. Nevertheless, the NPUs are constrained by similar memory bandwidths with the GPUs, thereby delivering comparable peak decoding speeds. A limitation of deploying on NPUs is the requisite extensive modification of LLM models to accommodate new model structures, demanding considerable time and effort. Therefore, deploying on GPUs presents a proficient complement to NPUs, supporting new models in both a swift and highly efficient manner.

## 4.2 Dynamic shape inference

Dynamic shape inference presents challenges related to shape expression, operator auto-generation, auto-optimization, memory reuse, and synchronization between CPU and GPU, etc. TVM and ONNX address these issues by utilizing symbols to represent unknown shape dimensions, allowing TVM to deeply optimize operators' implementation by understanding the shape relationships between tensors. Approaches such as Nimble[31] and DISC[43] have proposed dedicated Intermediate Representations (IR) to facilitate the expression and derivation of dynamic shapes, as well as code generation for dynamic shape operators and runtime execution for dynamic shape models. Solutions like BladeDISC[41] invest in extensive operator fusions for dynamic shape models.

In this paper, we neither introduce new IR or runtime designations nor focus on the auto-generation or auto-optimization of dynamic shape operators' kernels. As ONNX already supports the use of strings as dynamic shape dimensions, we use symbolic expressions to accommodate comprehensive dynamic shape inference. Subsequently, we leverage the symbolic expression relationships between tensors to enable static memory reuse and optimizations, including operator fusion. Apart from operator fusion, we manually implement the majority of operators to ensure they are both highly efficient and support dynamic shape overload. Additionally, we provide an execution schedule based on operators' classification without requiring specialized runtime designs.

## 4.3 4bits quantization

4-bit group-wise quantization is widely utilized for quantizing LLM, as the accuracy loss is generally acceptable[12, 22] and provides an optimal balance between accuracy and model size[9]. Our goal is not to introduce a novel quantization technique, such as GPTQ[12] or AWQ[22], but to present a new data storage format. Our E0M4 quantization approach can be seamlessly integrated with existing quantization techniques to minimize dequantization performance overhead while also slightly enhancing quantization accuracy.

LLM-FP4[23] and ZeroQuant-FP[34] also employ FP4 quantization and bit-shifting to accelerate data type converting. However, there is no standard definition for the allocation of exponential and fraction bits. Our method distinguishes itself by using all bits to store fraction bits, hence we refer to it as E0M4. The E0M4 FP4 format enables the conversion of FP4 to half-precision with just two bitwise operations.

## 4.4 KV cache optimizations

Several studies have quantized the KV cache to minimize memory consumption and allow the processing of longer token sequences by LLM[10, 13, 16, 37]. The PagedAttention[19] manages the KV cache through virtual memory with paging. Primarily, paged attention is deployed in server settings, and its feasibility for LLM deployment on device GPUs remains unexplored. Currently, we have not yet introduced any types of KV cache compression. In contrast, we have altered the KV cache format such that newly created KV caches can be directly appended to the end of old caches. We employ a sub-tensor-based method to eliminate the need for KV cache copy after each LLM inference iteration.

# 5 Conclusion

In this study, we propose several optimization techniques to facilitate high-efficiency LLM deployment on device GPUs: (a) a symbolic expression-based approach enabling dynamic shape models to be deployed on device GPUs; (b) operator optimizations and execution priority configurations to improve speed and reduce phone lagging; (c) the E0M4 FP4 quantization to minimize dequantization overhead; (d) KV cache copy optimizations to eliminate memory copy overhead after LLM inference.

With the integration of these optimizations, our Transformer-Lite engine achieves over 10x faster prefill speed and 2~3 times faster decoding speed compared with GPU-based MLC-LLM and CPU-based FastLLM. Specifically, we achieve prefill and decoding speeds of 330 token/s and 30 token/s for Gemma 2B, 121 token/s and 14 token/s for ChatGLM2 6B, respectively. Furthermore, we demonstrate deployment of Qwen1.5 14B based on a 24GB phone and achieve 54 token/s and 5 token/s for prefill and decoding speed respectively.

At present, the prefill speed of our engine is impeded by suboptimal matrix multiplication implementations. To enhance the prefill speed further, it's crucial to investigate more efficient matrix multiplication implementations, such as improving cache hit ratio and parallelism. Meanwhile, the decoding speed is constrained by memory bandwidth and attention computation at short and long prompt lengths respectively. Therefore, incorporating optimizations such as flash decoding, sparsity, and speculative decoding can improve the decoding speed.

# References

[1] Tom Brown, Benjamin Mann, Nick Ryder, Melanie Subbiah, Jared D. Kaplan, Prafulla Dhariwal, Arvind Neelakantan, Pranav Shyam, Girish Sastry, and Amanda Askell. 2020. Language models are few-shot learners. *Advances in neural information processing systems* 33, (2020), 1877–1901.

[2] Han Cai, Ji Lin, Yujun Lin, Zhijian Liu, Haotian Tang, Hanrui Wang, Ligeng Zhu, and Song Han. 2022. Enable deep learning on mobile devices: Methods, systems, and applications. *ACM Transactions on Design Automation of Electronic Systems (TODAES)* 27, 3 (2022), 1–50.

[3] Tianle Cai, Yuhong Li, Zhengyang Geng, Hongwu Peng, Jason D. Lee, Deming Chen, and Tri Dao. 2024. Medusa: Simple llm inference acceleration framework with multiple decoding heads. *arXiv preprint arXiv:2401.10774* (2024).

[4] Tianqi Chen, Thierry Moreau, Ziheng Jiang, Lianmin Zheng, Eddie Yan, Haichen Shen, Meghan Cowan, Leyuan Wang, Yuwei Hu, and Luis Ceze. 2018. {TVM}: An automated {End-to-End} optimizing compiler for deep learning. In *13th USENIX Symposium on Operating Systems Design and Implementation (OSDI 18)*, 2018. 578–594.

[5] Tri Dao. 2023. Flashattention-2: Faster attention with better parallelism and work partitioning. *arXiv preprint arXiv:2307.08691* (2023).

[6] Tri Dao, Dan Fu, Stefano Ermon, Atri Rudra, and Christopher Ré. 2022. Flashattention: Fast and memory-efficient exact attention with io-awareness. *Advances in Neural Information Processing Systems* 35, (2022), 16344–16359.

[7] Tri Dao, Daniel Haziza, Francisco Massa, and Grigory Sizov. 2023. *Flash-decoding for long-context inference*.

[8] Yunbin Deng. 2019. Deep learning on mobile devices: a review. In *Mobile Multimedia/Image Processing, Security, and Applications 2019*, 2019. SPIE, 52–66.

[9] Tim Dettmers and Luke Zettlemoyer. The case for 4-bit precision: k-bit inference scaling laws, 2022. *URL https://arxiv. org/abs/2212.09720*.

[10] Shichen Dong, Wen Cheng, Jiayu Qin, and Wei Wang. 2024. QAQ: Quality Adaptive Quantization for LLM KV Cache. *arXiv preprint arXiv:2403.04643* (2024).

[11] Elias Frantar and Dan Alistarh. 2023. Sparsegpt: Massive language models can be accurately pruned in one-shot. In *International Conference on Machine Learning*, 2023. PMLR, 10323–10337.

[12] Elias Frantar, Saleh Ashkboos, Torsten Hoefler, and Dan Alistarh. 2022. Gptq: Accurate post-training quantization for generative pre-trained transformers. *arXiv preprint arXiv:2210.17323* (2022).

[13] Suyu Ge, Yunan Zhang, Liyuan Liu, Minjia Zhang, Jiawei Han, and Jianfeng Gao. 2023. Model tells you what to discard: Adaptive kv cache compression for llms. *arXiv preprint arXiv:2310.01801* (2023).

[14] Albert Gu and Tri Dao. 2023. Mamba: Linear-time sequence modeling with selective state spaces. *arXiv preprint arXiv:2312.00752* (2023).

[15] Kaiming He, Xiangyu Zhang, Shaoqing Ren, and Jian Sun. 2016. Deep residual learning for image recognition. In *Proceedings of the IEEE conference on computer vision and pattern recognition*, 2016. 770–778.

[16] Coleman Hooper, Sehoon Kim, Hiva Mohammadzadeh, Michael W. Mahoney, Yakun Sophia Shao, Kurt Keutzer, and Amir Gholami. 2024. KVQuant: Towards 10 Million Context Length LLM Inference with KV Cache Quantization. *arXiv preprint arXiv:2401.18079* (2024).

[17] Andrew G. Howard, Menglong Zhu, Bo Chen, Dmitry Kalenichenko, Weijun Wang, Tobias Weyand, Marco Andreetto, and Hartwig Adam. 2017. Mobilenets: Efficient convolutional neural networks for mobile vision applications. *arXiv preprint arXiv:1704.04861* (2017).

[18] Xiaotang Jiang, Huan Wang, Yiliu Chen, Ziqi Wu, Lichuan Wang, Bin Zou, Yafeng Yang, Zongyang Cui, Yu Cai, and Tianhang Yu. 2020. MNN: A universal and efficient inference engine. *Proceedings of Machine Learning and Systems* 2, (2020), 1–13.

[19] Woosuk Kwon, Zhuohan Li, Siyuan Zhuang, Ying Sheng, Lianmin Zheng, Cody Hao Yu, Joseph Gonzalez, Hao Zhang, and Ion Stoica. 2023. Efficient memory management for large language model serving with pagedattention. In *Proceedings of the 29th Symposium on Operating Systems Principles*, 2023. 611–626.

[20] Teven Le Scao, Angela Fan, Christopher Akiki, Ellie Pavlick, Suzana Ilić, Daniel Hesslow, Roman Castagné, Alexandra Sasha Luccioni, François Yvon, and Matthias Gallé. 2022. Bloom: A 176b-parameter open-access multilingual language model. (2022).

[21] Yuhui Li, Fangyun Wei, Chao Zhang, and Hongyang Zhang. 2024. Eagle: Speculative sampling requires rethinking feature uncertainty. *arXiv preprint arXiv:2401.15077* (2024).

[22] Ji Lin, Jiaming Tang, Haotian Tang, Shang Yang, Xingyu Dang, and Song Han. 2023. Awq: Activation-aware weight quantization for llm compression and acceleration. *arXiv preprint arXiv:2306.00978* (2023).

[23] Shih-yang Liu, Zechun Liu, Xijie Huang, Pingcheng Dong, and Kwang-Ting Cheng. 2023. Llm-fp4: 4-bit floating-point quantized transformers. *arXiv preprint arXiv:2310.16836* (2023).

[24] Zechun Liu, Changsheng Zhao, Forrest Iandola, Chen Lai, Yuandong Tian, Igor Fedorov, Yunyang Xiong, Ernie Chang, Yangyang Shi, and Raghuraman Krishnamoorthi. 2024. MobileLLM: Optimizing Sub-billion Parameter Language Models for On-Device Use

Cases. *arXiv preprint arXiv:2402.14905* (2024).

[25] Zichang Liu, Jue Wang, Tri Dao, Tianyi Zhou, Binhang Yuan, Zhao Song, Anshumali Shrivastava, Ce Zhang, Yuandong Tian, and Christopher Re. 2023. Deja vu: Contextual sparsity for efficient llms at inference time. In *International Conference on Machine Learning*, 2023. PMLR, 22137–22176.

[26] Xupeng Miao, Gabriele Oliaro, Zhihao Zhang, Xinhao Cheng, Hongyi Jin, Tianqi Chen, and Zhihao Jia. 2023. Towards efficient generative large language model serving: A survey from algorithms to systems. *arXiv preprint arXiv:2312.15234* (2023).

[27] Xupeng Miao, Gabriele Oliaro, Zhihao Zhang, Xinhao Cheng, Zeyu Wang, Rae Ying Yee Wong, Zhuoming Chen, Daiyaan Arfeen, Reyna Abhyankar, and Zhihao Jia. 2023. Specinfer: Accelerating generative llm serving with speculative inference and token tree verification. *arXiv preprint arXiv:2305.09781* (2023).

[28] Iman Mirzadeh, Keivan Alizadeh, Sachin Mehta, Carlo C. Del Mundo, Oncel Tuzel, Golnoosh Samei, Mohammad Rastegari, and Mehrdad Farajtabar. 2023. Relu strikes back: Exploiting activation sparsity in large language models. *arXiv preprint arXiv:2310.04564* (2023).

[29] Bo Peng, Eric Alcaide, Quentin Anthony, Alon Albalak, Samuel Arcadinho, Huanqi Cao, Xin Cheng, Michael Chung, Matteo Grella, and Kranthi Kiran GV. 2023. Rwkv: Reinventing rnns for the transformer era. *arXiv preprint arXiv:2305.13048* (2023).

[30] Alec Radford, Jong Wook Kim, Chris Hallacy, Aditya Ramesh, Gabriel Goh, Sandhini Agarwal, Girish Sastry, Amanda Askell, Pamela Mishkin, and Jack Clark. 2021. Learning transferable visual models from natural language supervision. In *International conference on machine learning*, 2021. PMLR, 8748–8763.

[31] Haichen Shen, Jared Roesch, Zhi Chen, Wei Chen, Yong Wu, Mu Li, Vin Sharma, Zachary Tatlock, and Yida Wang. 2021. Nimble: Efficiently compiling dynamic neural networks for model inference. *Proceedings of Machine Learning and Systems* 3, (2021), 208–222.

[32] Hugo Touvron, Thibaut Lavril, Gautier Izacard, Xavier Martinet, Marie-Anne Lachaux, Timothée Lacroix, Baptiste Rozière, Naman Goyal, Eric Hambro, and Faisal Azhar. 2023. Llama: Open and efficient foundation language models. *arXiv preprint arXiv:2302.13971* (2023).

[33] Ji Wang, Bokai Cao, Philip Yu, Lichao Sun, Weidong Bao, and Xiaomin Zhu. 2018. Deep learning towards mobile applications. In *2018 IEEE 38th International Conference on Distributed Computing Systems (ICDCS)*, 2018. IEEE, 1385–1393.

[34] Xiaoxia Wu, Zhewei Yao, and Yuxiong He. 2023. Zeroquant-fp: A leap forward in llms post-training w4a8 quantization using floating-point formats. *arXiv preprint arXiv:2307.09782* (2023).

[35] Haojun Xia, Zhen Zheng, Yuchao Li, Donglin Zhuang, Zhongzhu Zhou, Xiafei Qiu, Yong Li, Wei Lin, and Shuaiwen Leon Song. 2023. Flash-llm: Enabling cost-effective and highly-efficient large generative model inference with unstructured sparsity. *arXiv preprint arXiv:2309.10285* (2023).

[36] Heming Xia, Tao Ge, Peiyi Wang, Si-Qing Chen, Furu Wei, and Zhifang Sui. 2023. Speculative decoding: Exploiting speculative execution for accelerating seq2seq generation. In *Findings of the Association for Computational Linguistics: EMNLP 2023*, 2023. 3909–3925.

[37] Wangsong Yin, Mengwei Xu, Yuanchun Li, and Xuanzhe Liu. 2024. LLM as a System Service on Mobile Devices. *arXiv preprint arXiv:2403.11805* (2024).

[38] Zhihang Yuan, Yuzhang Shang, Yang Zhou, Zhen Dong, Chenhao Xue, Bingzhe Wu, Zhikai Li, Qingyi Gu, Yong Jae Lee, and Yan Yan. 2024. LLM Inference Unveiled: Survey and Roofline Model Insights. *arXiv preprint arXiv:2402.16363* (2024).

[39] Susan Zhang, Stephen Roller, Naman Goyal, Mikel Artetxe, Moya Chen, Shuohui Chen, Christopher Dewan, Mona Diab, Xian Li, and Xi Victoria Lin. 2022. Opt: Open pre-trained transformer language models. *arXiv preprint arXiv:2205.01068* (2022).

[40] Yao Zhao, Zhitian Xie, Chenyi Zhuang, and Jinjie Gu. 2023. Lookahead: An Inference Acceleration Framework for Large Language Model with Lossless Generation Accuracy. *arXiv preprint arXiv:2312.12728* (2023).

[41] Zhen Zheng, Zaifeng Pan, Dalin Wang, Kai Zhu, Wenyi Zhao, Tianyou Guo, Xiafei Qiu, Minmin Sun, Junjie Bai, and Feng Zhang. 2023. Bladedisc: Optimizing dynamic shape machine learning workloads via compiler approach. *Proceedings of the ACM on Management of Data* 1, 3 (2023), 1–29.

[42] Yinmin Zhong, Shengyu Liu, Junda Chen, Jianbo Hu, Yibo Zhu, Xuanzhe Liu, Xin Jin, and Hao Zhang. 2024. DistServe: Disaggregating Prefill and Decoding for Goodput-optimized Large Language Model Serving. *arXiv preprint arXiv:2401.09670* (2024).

[43] Kai Zhu, W. Y. Zhao, Zhen Zheng, T. Y. Guo, P. Z. Zhao, J. J. Bai, Jun Yang, X. Y. Liu, L. S. Diao, and Wei Lin. 2021. DISC: A dynamic shape compiler for machine learning workloads. In *Proceedings of the 1st Workshop on Machine Learning and Systems*, 2021. 89–95.

[44] Democratizing on-device generative AI with sub-10 billion parameter models. Retrieved from https://www.qualcomm.com/news/onq/2023/09/democratizing-on-device-generative-ai-with-sub-10-billion-parameter-models

[45] Large Language Models On-Device with MediaPipe and TensorFlow Lite. Retrieved from https://developers.googleblog.com/2024/03/running-large-language-models-on-device-with-mediapipe-andtensorflow-lite.html

[46] MediaTek Demonstrating On-Device Generative AI Using Llama 2 LLM at MWC 2024. Retrieved from https://www.mediatek.com/blog/mediatek-dimensity-demos-on-device-generative-ai-using-meta-llama-2-llm

[47] Unlocking on-device generative AI with an NPU and heterogeneous computing. Retrieved from https://www.qualcomm.com/content/dam/qcomm-martech/dm-assets/documents/Unlocking-on-device-generative-AI-with-an-NPU-and-heterogeneous-computing.pdf

[48] TensorFlow Lite. Retrieved from https://tensorflow.google.cn/lite

[49] NCNN. Retrieved from https://github.com/Tencent/ncnn

[50] llama.cpp. Retrieved from https://github.com/ggerganov/llama.cpp

[51] MLC-LLM. Retrieved from https://github.com/mlc-ai/mlc-llm

[52] fastllm. Retrieved from https://github.com/ztxz16/fastllm

[53] ONNX. Retrieved from https://github.com/onnx/onnx

[54] SymPy. Retrieved from https://www.sympy.org/en/index.html

[55] clCreateImage. Retrieved from https://registry.khronos.org/OpenCL/sdk/3.0/docs/man/html/clCreateImage.html

[56] cl_qcom_priority_hint. Retrieved from https://github.com/willhua/QualcommOpenCLSDKNote/blob/master/docs/extensions/cl_qcom_priority_hint.txt

[57] cl_khr_priority_hints. Retrieved from https://registry.khronos.org/OpenCL/sdk/3.0/docs/man/html/cl_khr_priority_hints.html

[58] IEEE-754. Retrieved from https://en.wikipedia.org/wiki/IEEE_754

[59] Half-precision floating-point format. Retrieved from https://en.wikipedia.org/wiki/Half-precision_floating-point_format

[60] export_llama_to_onnx. Retrieved from https://github.com/luchangli03/export_llama_to_onnx

[61] ArchProbe. Retrieved from https://github.com/microsoft/ArchProbe

[62] clip-vit-base-patch32. Retrieved from https://huggingface.co/openai/clip-vit-base-patch32